%% file: neurips_2026.tex
\documentclass{article}

 \usepackage[preprint]{neurips_2026}

\usepackage[utf8]{inputenc} 
\usepackage[T1]{fontenc}    
\usepackage{hyperref}       
\usepackage{url}            
\usepackage{booktabs}       
\usepackage{amsfonts}       
\usepackage{nicefrac}       
\usepackage{microtype}      
\usepackage{xcolor}         
\usepackage{graphicx}
\usepackage{amsmath}
\usepackage{multirow}
\usepackage[T1]{fontenc}
\usepackage{xspace}
\usepackage[dvipsnames]{xcolor}
\usepackage{wrapfig}
\usepackage{cleveref}
\usepackage{tcolorbox}
\usepackage{xcolor}
\definecolor{emreblue}{HTML}{4d7ea8}
\NewDocumentCommand{\emre}
{ mO{} }{\textcolor{emreblue}{\textsuperscript{\textit{Emre}}\textsf{\textbf{\small[#1]}}}}

\usepackage{xcolor}
\newcommand{\pos}[1]{{\scriptsize\textcolor{green!50!black}{$\uparrow$#1}}}
\newcommand{\negi}[1]{{\scriptsize\textcolor{red!70!black}{$\downarrow$#1}}}

\title{{\fontfamily{lmdh}\selectfont AcquisitionSynthesis}: Targeted Data Generation using Acquisition Functions}

\author{
Ishika Agarwal*$^1$, Sofia Stoica*$^1$, Emre Can Acikgoz$^1$, \\ \textbf{Pradeep Natarajan$^2$, Mahdi Namazifar$^2$, Jiaqi Ma$^1$, Dilek Hakkani-Tür$^1$}\\
\vspace{2pt}
University of Illinois Urbana-Champaign$^1$, Amazon$^2$\\
\texttt{Correspondence to \{ishikaa2, sstoica2\}@illinois.edu}\\
}

\begin{document}

\maketitle

\newcommand{\sysn}{{\fontfamily{lmdh}\selectfont AcquisitionSynthesis}\xspace}
\newcommand{\sysnshort}{{\fontfamily{lmdh}\selectfont AS}\xspace}

\definecolor{takeawayorange}{HTML}{EF8600}


\begin{abstract}
Data quality remains a critical bottleneck in developing capable, competitive models. Researchers have explored many ways to generate top quality samples. Some works rely on rejection sampling: generating lots of synthetic samples and filtering out low-quality samples. Other works rely on larger or closed-source models to extract model weaknesses, necessary skills, or a curriculum off of which to base data generation. These works have one common limitation: there is no quantitative approach to measure the impact of the generated samples on the downstream learner. Active learning literature provides exactly this, in the form of acquisition functions. Acquisition functions measure the informativeness and/or influence of data, providing interpretable, model-centric signals. Inspired by this, we propose \sysn: using acquisition functions as reward models to train language models to generate higher-quality synthetic data. We conduct experiments on classic verifiable tasks of math, medical question-answering, and coding. Our experimental results indicate that (1) student models trained with \sysn data achieve good performance on in-distribution tasks (2-7\% gain) and is more robust to catastrophic forgetting, and (2) \sysn models can generate data for other models and for low-to-high resource training paradigms. By leveraging acquisition rewards, we seek to demonstrate a principled path toward model-aware self-improvement that surpasses static datasets.
  
\end{abstract}

\section{Introduction}

Despite data quality being a crucial driver to Large Language Model (LLM) progress, the guidelines for curating high quality data are not standardized or clear \citep{mehri2025samplelevelfeedbackusingreferencelevel, influence, li2025datacomplmsearchgenerationtraining, djuhera2025fixingpostcomparativestudy, kim-etal-2025-evaluating}. The current literature on data curation for LLMs is structured in two: (1) data selection, involving sampling a subset of data from a large set of given training data, and (2) synthetic data generation, involving generating data using LLMs based on given guidelines (either a set of skills/weaknesses/feedback \citep{sun2025codeevo, he2026stat, mehri2025samplelevelfeedbackusingreferencelevel} or a seed set of training data \citep{sudalairaj2024lab, jung2025prismaticsynthesisgradientbaseddata, cheng-etal-2025-star}). Our work focuses on \textbf{synthetic data generation}.

In this field, there are two general focuses: ensuring (1) \textbf{high quality data}, which requires generating large pools of synthetic data and applying rejection sampling to filter out low-quality samples \citep{selfinstruct, jung2025prismaticsynthesisgradientbaseddata, ge2025scalingsyntheticdatacreation}, (2) \textbf{model-centric data}, which relies on languages models (within multi-step pipelines) to identify model weaknesses and construct training curricula \citep{dataenvgym, liang2025swsselfawareweaknessdrivenproblem}. These methods can use a lot of compute to generate-then-filter, and are not always interpretable because using a qualitative approach that cannot measure the downstream impact of generated data provides no justification for why a given sample is useful. More importantly, despite the successes of these works, a fundamental question still remains unexplored: 

\begin{quote}
    \textit{Can simple, principled notions of data utility guide LLMs to generate better data?}
\end{quote}

The idea of data utility has been studied extensively in active learning \citep{settles2012active}. 



Active learning, which involves selecting the most informative unlabeled samples for oracle annotation to reduce labeling cost while maximizing learning signal \citep{settles2012active}, has helped define simple and robust methods for measuring data utility called \textit{acquisition functions}. In our work, we introduce \sysn -- a framework to train our models using Group Relative Policy Optimization (GRPO) \cite{shao2024deepseekmathpushinglimitsmathematical} with rewards inspired by acquisition functions to generate directly-usable, high quality data.

\begin{figure}
    \centering
    \includegraphics[width=0.49\linewidth]{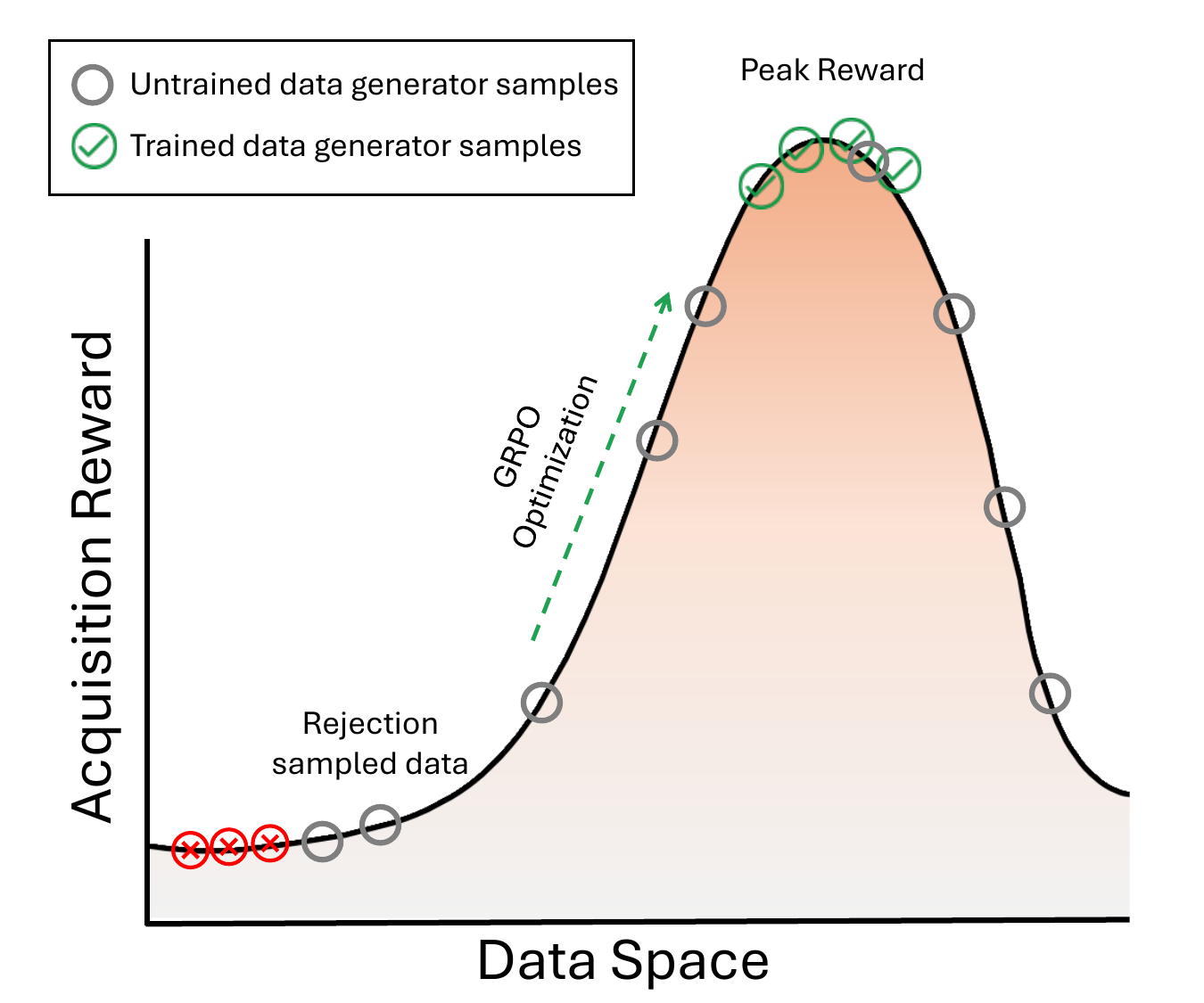}
    \includegraphics[width=0.49\linewidth]{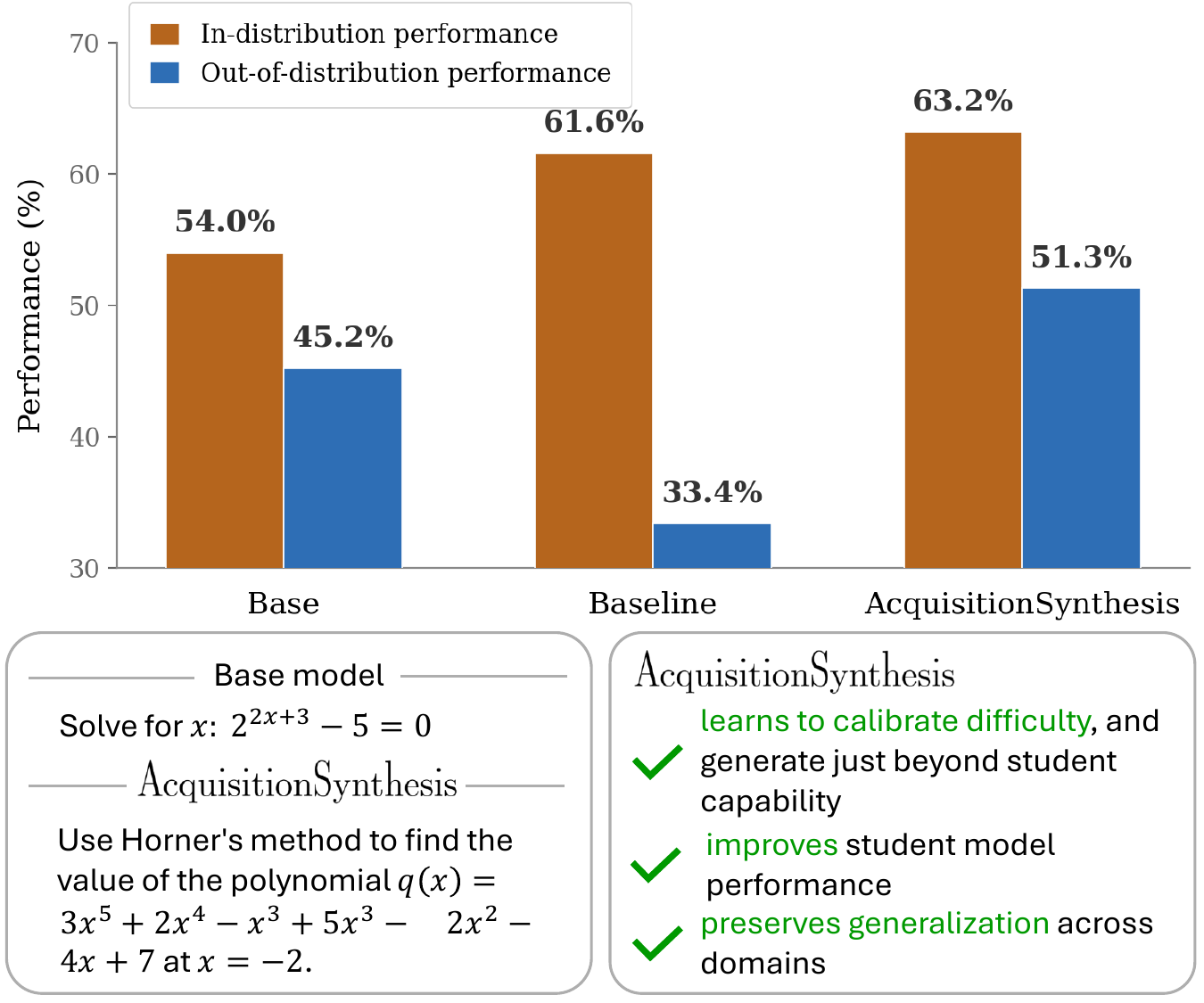}
    \caption{\textbf{Left:} \sysn shifts generated samples toward high acquisition reward regions of the data space. Grey circles: samples from an untrained generator, scattered with varying rewards. Red ×'s: samples rejected by rejection sampling due to low reward. Green checks: samples from a trained \sysn generator, concentrated near peak reward. \textbf{Right:} Plotted is the average in-distribution and out-of-distribution performance across tasks. While baselines improve in-distribution performance at the cost of OOD generalization, \sysn improves both.}
    \label{fig: acquisition_synthesis}
\end{figure}

Figure \ref{fig: acquisition_synthesis}(Left) illustrates our central finding: acquisition functions, when used as RL rewards, are simple yet surprisingly effective. By optimizing for these rewards, the generator learns to directly produce samples that are informative for a student model\footnote{Unless stated otherwise, the student model is usually an instantiation of the \sysn model's untrained version.}. Figure \ref{fig: acquisition_synthesis}(Right) overviews the effects of \sysn: acquisition rewards can be used to assess a student model's weaknesses and adjust the data generation to push the student's knowledge just beyond what it already knows.

\paragraph{Overview}
In \sysn, to define data utility, we use five complementary acquisition rewards motivated by active learning literature that capture different dimensions of data quality: Confidence \citep{dredze2008active, qi2026generationactivelearningmixture}, Proximity \citep{less}, Gradient Magnitude \citep{ash2020deepbatchactivelearning, gul2024lplgradoptimizingactivelearning}, Diversity \citep{wang2017uncertainty, jung2025prismaticsynthesisgradientbaseddata}, and Answer Variance \citep{diao2024activepromptingchainofthoughtlarge, kee2018query, cao2021bayesianactivelearningdisagreements}. We test \sysn on verifiable tasks such as math, medical QA, and coding; by using \sysn models to construct a dataset, we can train a student model on that dataset to determine the quality of the data generator (see Figure \ref{fig: method}). We release our code \href{https://anonymous.4open.science/r/AcquisitionSynthesis-EAE7/}{here}.

Our analyses uncover the following:
\begin{enumerate}
    \item Student models trained on \sysn data improve by $\sim$2-7\%, with higher gains for bigger models (Section \ref{sec: main_results}).
    \item Student models trained with \sysn suffer less from catastrophic forgetting compared to baselines, and can even improve performance in out-of-distribution settings by ~3\% (Section \ref{sec: domain_generalization}).
    \item For larger data-generator models, \sysn trains stronger generators that generalize across model families and sizes (Section \ref{sec: student_generalization}).
    \item \sysn generates data suitable for various training paradigms (Section \ref{sec: training_paradigm}).
\end{enumerate}

\section{Related Work}
\paragraph{Learning Signals.} Quantifying what makes a training sample ``useful'' has a long history in active learning \citep{settles2012active}, where \textit{acquisition functions} guide the selection of which unlabeled samples to annotate. These acquisition functions use signals from geometry-based heuristics \citep{kang2025pcoreseteffectiveactivelearning, sener2018activelearningconvolutionalneural, mirzasoleiman2020coresetsdataefficienttrainingmachine} to uncertainty or margin-based sampling \citep{balcan2007margin, lakshminarayanan2017simplescalablepredictiveuncertainty}. Recent work has adapted these ideas to LLMs: \citep{agarwal2025deliftdataefficientlanguage, rouzegar-makrehchi-2024-enhancing} use model uncertainty, \citep{jung2025prismaticsynthesisgradientbaseddata, zhang2026syntheticdatagenerationtraining} design diversity-based sampling methods, and \cite{Bayer_2026} employ LLM-as-a-Judge scores as a quality proxy. However, all of these works operate in the \textit{data selection} realm, where there exists a large pool of data already. In our paper, we build upon these works to show that language models can optimize data generation when given such a data quality reward.

\paragraph{Synthetic Data Generation.} There are a variety of use cases that require models to generate their own training data: when there isn't enough data, when the data has to be targeted for a particular domain, or the data has to adhere to certain stylistic guidelines. Self-Instruct \citep{selfinstruct} was one of the first papers that showed that language models can generate coherent samples and, indeed, improve language model capability. Since then, the focus of data generation has shifted to targeted data generation, focusing on (1) complexity \citep{xu2025wizardlmempoweringlargepretrained}, (2) persona-driven diversity \citep{ge2025scalingsyntheticdatacreation}, and (3) addressing model weaknesses \citep{dataenvgym}. We reformulate data generation as an optimization problem (where GRPO \citep{shao2024deepseekmathpushinglimitsmathematical} is the optimization mechanism), in which the acquisition reward formulation is our core contribution: a set of principled, interpretable objectives (grounded in active learning theory) that give the generator a quantitative target for what constitutes useful training data.

\section{Methodology}
Motivated by the active learning literature, we identify five acquisition function formulations to act as reward functions. Each of them are adapted for language models/transformers. Figure \ref{fig: method} contains an illustration of our evaluation framework. The data generator (i.e., \sysn) model is trained using 500 samples --these samples are used as in-context learning examples (see Appendix \ref{app: prompts} for more details)-- during training, the \sysn model generates a data sample which is fed to the student model. To clarify, the \sysn model and student model are two instantiations of the same model. The acquisition reward is calculated with respect to the student model, in one of the ways below, and the reward is sent back to the \sysn model to update its parameters.

Here are the different acquisition functions we use:
\begin{enumerate}
    \item \textbf{Confidence}: this acquisition function rewards a model based on how \textit{un}-confident the student model is on a generated data point. In active learning literature, this is also known as "uncertainty sampling" or "margin sampling" \citep{balcan2007margin}. We calculate confidence by the average difference in probability between the top-2 most probable tokens over the sequence\footnote{Given a prompt $P$ and response $R$ (where $R_t^{(i)}$ is the $i$-th most probable token in the $t$-th spot), the LLM's confidence is: $conf(P, R) = \frac{1}{|R|} \sum_{t=1}^{|R|} \left( p_\theta(R_t^{(1)} \mid P, R_{<t}) - p_\theta(R_t^{(2).} \mid P, R_{<t}) \right)$}, and assign an acquisition reward based on the inverse of the average\footnote{Following $conf(P,R)$, the reward is $\frac{1}{conf(P,R)}$.}. This acquisition function incentivizes models to generate data points that they are most uncertain about \cite{dredze2008active, qi2026generationactivelearningmixture}, in order to learn most about the task. 
    
    \item \textbf{Proximity}: we reward the model based on how semantically similar the generated data point is to the given task's training data to avoid irrelevant samples \citep{less}. We cluster the data using HDBSCAN \citep{hdbscan}, a dynamic method of clustering that does not require specification of the number of clusters beforehand. The acquisition reward assigned is the cosine distance from the generated sample to the closest cluster center.
    
    \item \textbf{Gradient Magnitude}: this acquisition function assigns a reward based on how significant the learning signal is \citep{ash2020deepbatchactivelearning, gul2024lplgradoptimizingactivelearning}, i.e., the Frobenius norm of the gradient induced by backpropagating on the student model given the generated data sample. This incentivizes the model to generate samples that ``surprise'' the student model and can achieve the highest learning signal from.
    
    \item \textbf{Diversity}: we reward the model based on how semantically diverse the generated data point is to the 500 training samples to avoid mode collapse. This incentivizes the model to generate samples that are different from what it has already seen before \citep{wang2017uncertainty, jung2025prismaticsynthesisgradientbaseddata}. It is the opposite of the Proximity reward: $1-p$ where $p$ is the proximity reward.
    
    \item \textbf{Answer variance}: this acquisition function rewards a model based on the number of unique answers it outputs in $k=8$ model inferences. In active learning literature, this is close to the "query-by-committee" sampling strategy \citep{kee2018query, burbidge2007active}. We assign a higher reward if the student model is more confused. We cluster the responses using HDBSCAN, and report the number of clusters as the reward. This incentivizes the model to generate samples that the student model is uncertain how to answer \citep{diao2024activepromptingchainofthoughtlarge, kee2018query, cao2021bayesianactivelearningdisagreements}. The motivation is similar to Confidence; however, Confidence measures token-level uncertainty while Answer Variance measures semantic-level uncertainty. 
\end{enumerate}

To ensure that our generated data is parseable during post-processing, we apply a format reward following previous RLVR works. The reward is $0$ if the generated data is parseable (via HTML tags such as ``question'', ``reasoning'', and ``answer'') and $-1$ if there are any errors. This reward is applied during all \sysn training, but we also use this as an ablation in our experimentation, and call it \textbf{Format}.

\subsection{Dataset Construction for Student Model}
Since we ask the model to generate its own training labels, label reliability is an open research problem. Following prior work, we assign pseudo-labels based on agreement across multiple sampled outputs. In our open-ended tasks, where exact string matching is infeasible, we embed the $N$ sampled responses with a sentence encoder (\texttt{Qwen/Qwen3-Embedding-0.6B} \citep{qwenembedding}, in particular because of its small size yet strong performance on MTEB \citep{mteb}), cluster them by cosine similarity, and treat the medoid of the largest cluster as the pseudo-label. In Appendix \ref{sec: gpt_answer}, we run a small-scale experiment to see the effect of more reliable labeler (GPT-4o-mini) and see that we can get a 6\% performance boost if we use better labels.

\section{Experimental Evaluation}
\begin{figure}
    \centering
    \includegraphics[width=\linewidth]{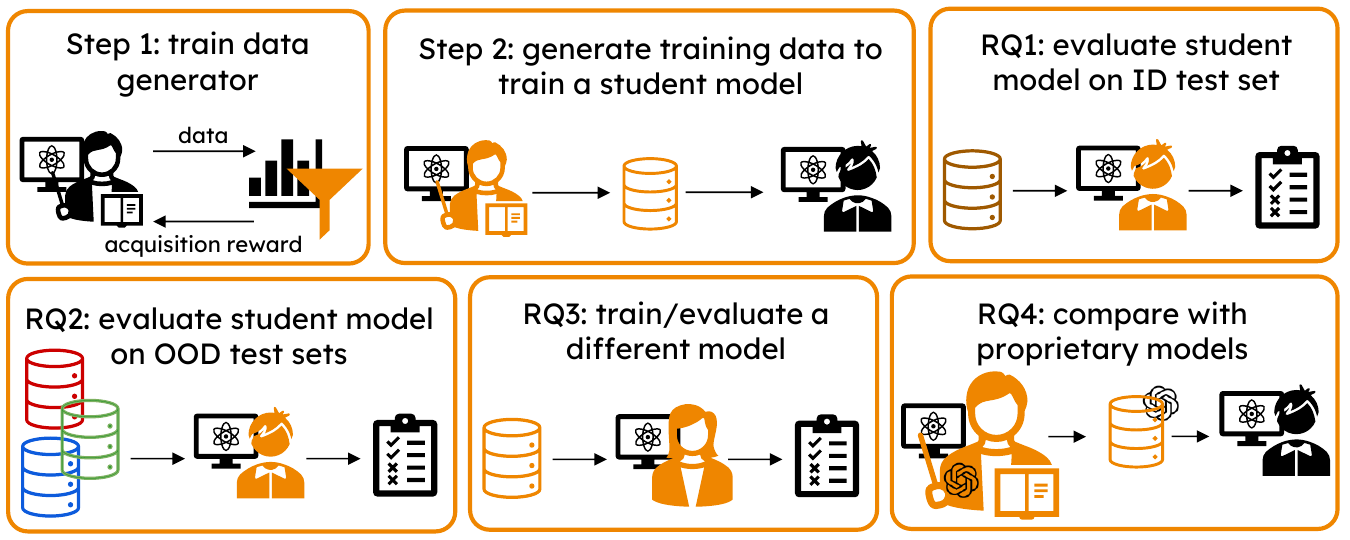}
    \caption{an illustration of our evaluation framework. Steps 1 and 2 describe our dataset generation while our evaluation for RQ's 1-4 are illustrated. More details are introduced in Section \ref{sec: results}. To clarify, we only use the generated dataset from \sysn to train our student model.}
    \label{fig: method}
\end{figure}

Figure \ref{fig: method} contains an illustration of our experimental setup. In Step 1, we train our \sysn model with 500 seed samples of data (again, as ICL examples -- see Appendix \ref{app: prompts}). Next, we use the trained model to generate training data for a student model (Step 2). We generate 1,000 samples (and keep this consistent with all of our baselines). After training the student model with SFT on the generated samples, we evaluate the student model on a test set. We aim to achieve better student model performance than using the original curated datasets or generated data from other methods. We use 6 A100 NVIDIA GPUs for performing GRPO training, which takes roughly 30-60 minutes for training \sysn models.

\paragraph{Models.} For main results, we use \texttt{Qwen2.5-3B-Instruct} (shorthand: Qwen 3B) and \texttt{Llama-3.1-8B-Instruct} (shorthand Llama 8B). To show the generalization of our method to various model sizes and architectures, we also use \texttt{Qwen2.5-7B-Instruct} (shorthand: Qwen 7B).

\paragraph{Datasets.} We evaluate on three types of datasets: math, medical, and coding. For math, we train on Numina (\texttt{AI-MO/NuminaMath-CoT}) \citep{numina} and evaluate on a held-out portion of Numina as well as AIME24 (\texttt{HuggingFaceH4/aime\_2024}). For medical, we train on MedMCQA (\texttt{openlifescienceai/medmcqa}) \citep{medmcqa} and evaluate on a held-out test set of MedMCQA as well as PubMedQA (\texttt{qiaojin/PubMedQA}) \citep{pubmedqa}. Finally, for coding, we only use CodeForces (\texttt{open-r1/codeforces-cots}) \citep{codeforces} for evaluation.

\paragraph{Baselines}
We use the following data synthesis baselines to test the robustness of our method:
\begin{enumerate}
    \item \textbf{Original}: we use the frozen model to generate data samples, with no guidance.
    \item \textbf{DataEnvGym} \citep{khan2025dataenvgymdatagenerationagents}: given data for a particular task, a student model is evaluated on the data set and the samples on which it performed poorly are collected. A data generator model is then used to identify the skills required to solve those samples, and generate new question-answer pairs to target those weaknesses. Using this baseline, we aim to show the benefit of using more principled and amortized approaches for determining weaknesses and generating samples to target weaknesses.
    \item \textbf{Prismatic Synthesis} \citep{jung2025prismaticsynthesisgradientbaseddata}: this method curates diverse data by generating 100,000 data samples, clustering them by their gradient vectors, and discarding the samples that belong to dense gradient clusters. Using this baseline, we aim to show the benefits of removing post-hoc data filtering.
\end{enumerate}

We also evaluate on the following two data-selection baselines to show why synthesis can produce higher-quality samples than pruning existing training data:
\begin{enumerate}
    \item \textbf{Random}: we randomly select 1000 data samples to use as training data for the student model.
    \item \textbf{Filtered}: we use our acquisition functions in a pool-based active learning set up. We score each data sample in the pool using the acquisition functions, normalize the scores, and average across the five acquisition functions to get a final ranking. We prune out the top 1000 data samples to use as training data. 
\end{enumerate}

\section{Results}
\label{sec: results}
We break our results and analysis into four research questions (RQs) (see Figure \ref{fig: method} for illustrations of our RQ evaluations) regarding in-distribution vs out-of-distribution performance, student model transfer, and student model training paradigms. We also analyze the token efficiency of \sysn models. The details are in the following sections.

\subsection{RQ1: how well does \sysn generate data?}
\label{sec: main_results}
\input{results_tables/main_results}

Table \ref{tab: main_results} contains our results for in-distribution performance, showing the performance of all our baselines and \sysn variants, with Qwen 3B, Qwen 7B, and Llama 8B. On average, \sysn improves model performance by 1.82\% for Qwen and 7.47\% for Llama. The data selection baselines achieve meager improvements -- this might be because of the question difficulty; as the untrained model's performance improves, the gains due to data selection also improve. Hence, there are two key limitations of data selection: (1) a large pool of questions is assumed, and (2) question difficulty does not always correlate to high learning signal. On the other hand, \sysn and the other \textbf{data synthesis methods are able to calibrate the question difficulty} to ensure the model learns most effectively from its data.

Regarding the data synthesis methods, Original is an unreliable baseline as its performance is setting dependent: it is sometimes a strong baseline (Llama 8B for Numina), and other times it does not yield performance gains (Qwen 3B). DataEnvGym and Prismatic Synthesis are strong baselines -- they directly target model weaknesses and are able to improve a model's performance on the test set. Moreover, Prismatic Synthesis uses a small, proxy model to compute gradients in order to cluster and select diverse gradients. Following their setup, we use Qwen2.5-0.5B-Instruct, which could explain why Prismatic Synthesis works better on Qwen 3B than on Llama 8B. To conclude, \textcolor{OliveGreen}{models trained with \sysn data show competitive performance, no matter the model family or size.}

\input{results_tables/generalization_qwen}
\subsection{RQ2: how do models trained with \sysn data perform on OOD tasks?}
\label{sec: domain_generalization}

To measure for catastrophic forgetting, we evaluate each trained student on other out-of-distribution datasets: one in the same domain, and two in different domains. To clarify, if a student were trained on Numina (domain: math), the same domain dataset would be AIME and the different domain datasets would be MedMCQA and CodeForces. Tables \ref{tab: generalization_results_qwen} and \ref{tab: generalization_results_llama} contain those results.

Student models trained with the data selection baselines perform inconsistently in out-of-distribution settings, where Qwen 3B achieves a 2.56\% increase, while Llama 8B achieves a 5.11\% decrease in performance. Data selection assumes that the training distribution matches the testing distribution: our results indicate that when the distributions don't match, the performance drops in OOD settings. Regarding the data synthesis baselines, even though we saw good performance for in-distribution tasks, the same performance gains do not transfer OOD. Collectively, the synthesis baselines degrades Qwen's performance by 1.25\% and Llama's performance by 12.94\%. On the other hand, \sysn improves Qwen by 3.43\% and Llama maintains its average performance (only a slight increase of 0.25\%). These experiments show that \textcolor{OliveGreen}{student models trained on \sysn data are less prone to catastrophic forgetting, compared to baselines}.

\input{results_tables/generalization_llama}

\vspace{-3mm}
\newtcolorbox{takeaway1}{
  colback=takeawayorange!5,
  colframe=takeawayorange,
  title=Key Takeaway from RQ1 and RQ2,
  fonttitle=\bfseries\small,
  boxsep=2pt, left=4pt, right=4pt, top=2pt, bottom=2pt, toptitle=1pt, bottomtitle=1pt, before skip=8pt, after skip=0pt
}

\begin{takeaway1}
 \sysn generates data that is not only useful for in-domain learning, but also better aligned with the broader evaluation distribution. Furthermore, Confidence is the best general-purpose reward, and Diversity offers the best cost-to-OOD-gain tradeoff.
\end{takeaway1}

\paragraph{Why \sysn Generalizes.}
\begin{wraptable}{r}{0.6\textwidth}
\centering
\vspace{-3mm}
\resizebox{\linewidth}{!}{
\begin{tabular}{lcccc}
\toprule
\textbf{Model} & \textbf{AIME} & \textbf{MedMCQA} & \textbf{CodeForces} & \textbf{Avg.} \\
\midrule
DataEnvGym~\citep{dataenvgym} & 0.558 & 0.464 & 0.549 & 0.524 \\
Prismatic Syn.~\citep{jung2025prismaticsynthesisgradientbaseddata} & 0.566 & 0.456 & 0.557 & 0.526 \\
Confidence (Ours) & \textbf{0.575} & \textbf{0.497} & \textbf{0.560} & \textbf{0.544}  \\
\bottomrule
\end{tabular}
}
\caption{Distributional coverage of generated data on OOD benchmarks.}
\vspace{-5mm}
\label{tab:corr_comparison}
\end{wraptable}%

To explain the OOD gains of \sysn, we measure the distributional alignment between the generated training data and held-out benchmark data, computed as the average cosine similarity between their embeddings~\citep{acikgoz2026toolr0}. Higher values indicate that synthesis covers the regions of input space encountered at test time. \Cref{tab:corr_comparison} shows that our Confidence variant attains the highest alignment on every benchmark compared to previous works. This suggests that uncertainty-guided generation better covers the regions of the test distribution where the model is likely to make mistakes, rather than only producing diverse or weakness-targeted samples~\citep{huang2023llmsselfimprove}. These results provide one explanation for the stronger generalization observed in RQ2: \sysn generates data that is not only useful for in-domain learning, but also better aligned with the broader evaluation distribution. 

\vspace{-0.4cm}
\subsection{RQ3: can \sysn models generalize and generate data for other models?}
\label{sec: student_generalization}

\input{results_tables/model_transfer}

In production settings, we can not guarantee that the student model will have the same architecture or size as the generator model. Thus, this research question explores our trained \sysn models' ability to generate data for models of varying size and architectures. We train Qwen 3B, Qwen 7B, and \texttt{meta-llama/Llama-3.2-3B-Instruct} (Llama 3B) from data generated by Llama 8B. Table \ref{tab: model_transfer_results_llama} contains the results. We don't report results on the opposite setting (of training models with data generated by Qwen 3B) because Qwen 3B is too small of a model to distill data from -- the notion of improving weak-to-strong model generalization is out-of-scope for our paper.

The data synthesis baselines are not able to generate data for other models: Original generalizes well on Llama 3B and Qwen 3B because that is directly knowledge distillation. DataEnvGym performs badly because the generated data is supposed to target a particular model's weaknesses, and, thus, cannot generalize to other models. Prismatic Synthesis's assumption of diverse data is slightly inconsistent: good for some domains, but not all. Meanwhile, \textcolor{OliveGreen}{\sysn methods maintain stable improvement across model families and sizes} -- especially on datasets where untrained models do not perform well (in particular, math). 

\subsection{RQ4: how does student model performance vary when trained in low-to-high resource training methods?}
\label{sec: training_paradigm}

\begin{figure}[h]
    \centering
    \includegraphics[width=0.9\linewidth]{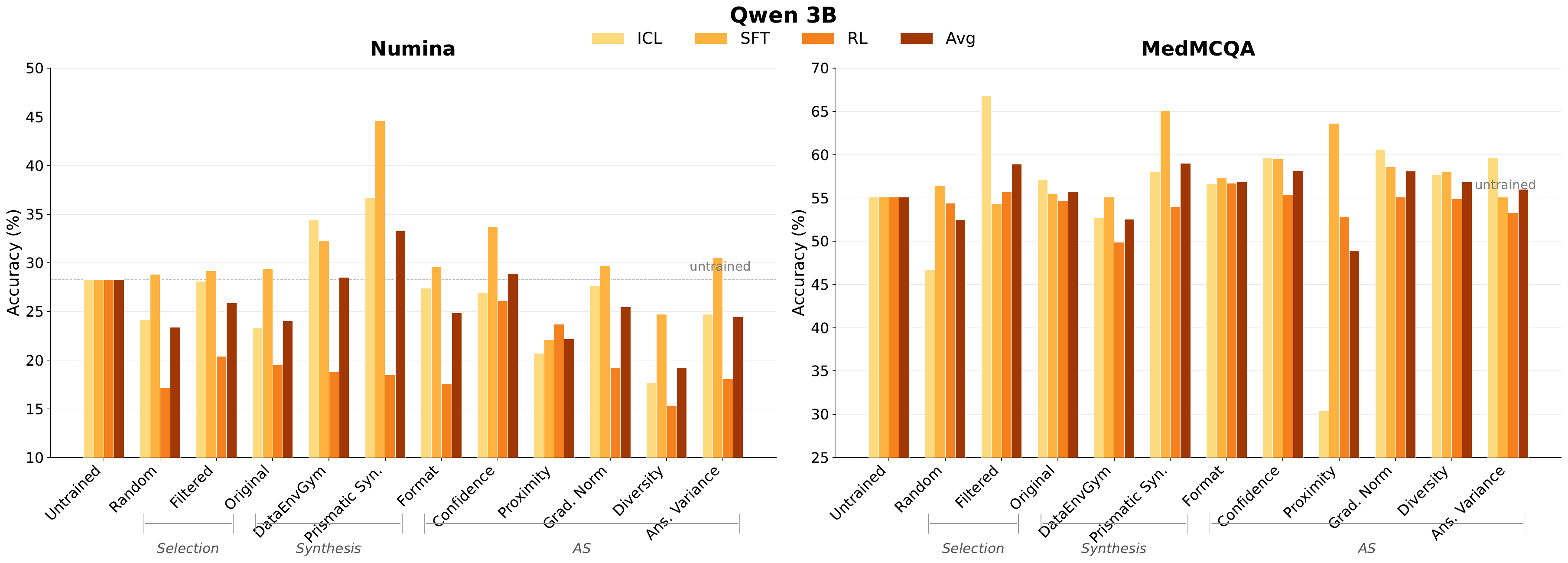}
    \includegraphics[width=0.9\linewidth]{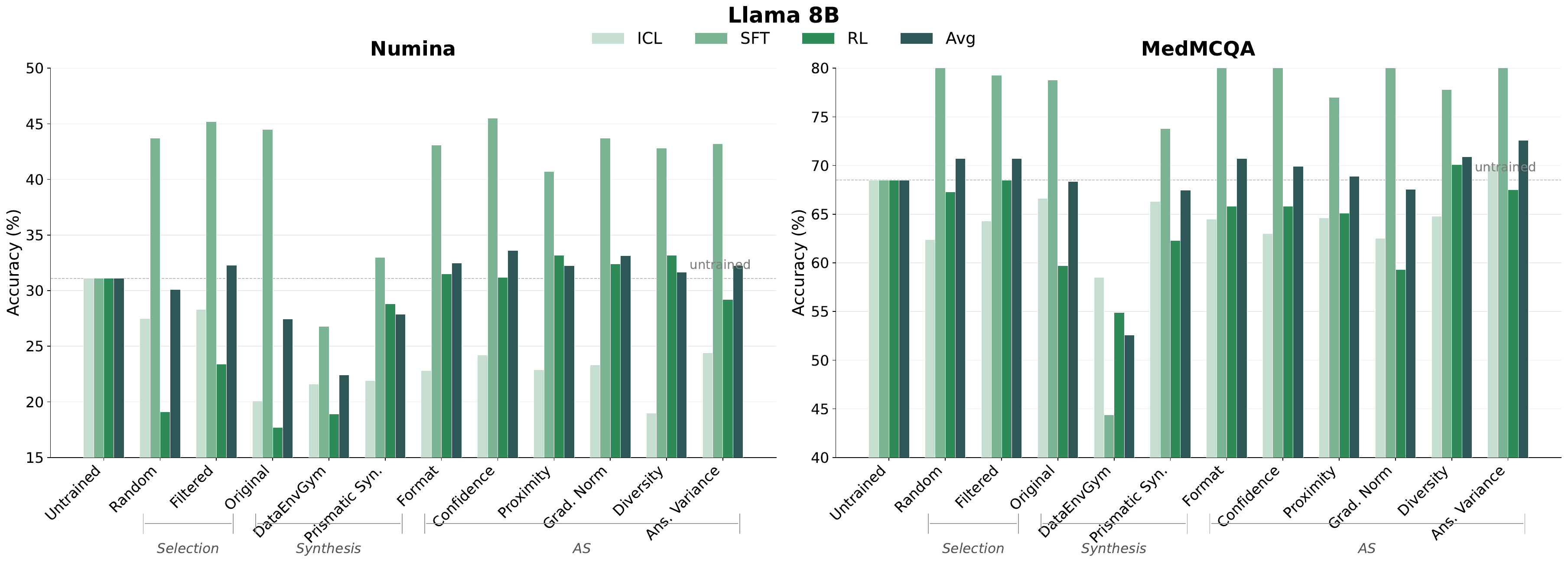}
    \caption{Performance of student models with varying training paradigms. We use the data selected/generated by each method and perform either ICL, SFT, or RL to obtain our student model.}
    \label{fig: student_result}
\end{figure}
In our work so far, we have used the curated (selected or generated) datasets as SFT datasets. But can we also use them in resource-constrained (in-context learning, or ICL) or resource-rich environments (reinforcement learning, or RL)? Figure \ref{fig: student_result} contains the results. Overall, we see SFT remains the strongest paradigm, but this is expected because (1) 3-billion parameter models might not be strong enough to use in-context examples effectively, and (2) we use an accuracy reward for RL training, which might be too sparse for a small model. For \sysn, the direction of performance gains is mostly consistent across training paradigms: if the student model improves with SFT, it also improves with ICL and RL. This consistency is not guaranteed for baselines -- for example, (1) Filtered improves substantially with ICL on MedMCQA but degrades with SFT, or (2) Prismatic Synthesis degrades with RL on Numina, despite strong SFT gains.

Our hypothesis for why \sysn can adapt to any training paradigm is that each training paradigm requires data with different signals (ICL might require data with the maximum teachable content, SFT would benefit from large, un-noisy gradient signals, and RL is most effective with data that has dense learning signals). Our baselines optimize for a single axis of this: Prismatic Synthesis optimizes for diversity which is why it does well SFT; Random optimizes for coverage; DataEnvGym optimizes for model weaknesses, which provides inconsistent signals (SFT is the best training paradigm for Qwen 3B, but the worst for Llama 8B). \sysn, on the other hand, provides one framework to generate data for any kind of training stage that necessitates any particular data characteristic. To summarize, \textcolor{OliveGreen}{\sysn data can be reliably used to train models in resource-constrained (ICL) to resource-abundant (RL) environments without restructuring the data}.

\newtcolorbox{takeaway2}{
  colback=takeawayorange!5,
  colframe=takeawayorange,
  title=Key Takeaway from RQ3 and RQ4,
  fonttitle=\bfseries\small,
  boxsep=2pt, left=4pt, right=4pt, top=2pt, bottom=2pt, toptitle=1pt, bottomtitle=1pt, before skip=8pt, after skip=0pt
}

\begin{takeaway2}
 Data generated from \sysn models can be used flexibly to train student models of different families and sizes, and for any training paradigm (ICL, SFT, RL) reliably.
\end{takeaway2}

\subsection{Discussion on token efficiency}
During the data generation phase, we only apply one filtering pass: we discard any exact-match, duplicate questions. Due to this filtering, on average, Original discards 26.0\% of generated data, DataEnvGym discards 76.7\%, Prismatic Synthesis 65.2\% (via its own rejection sampling), and \sysn discards 11.1\% of data. Overall, \textcolor{OliveGreen}{\sysn's reward design is able to efficiently generate a dataset by discarding 44.9\% fewer tokens}.

\vspace{-0.2cm}
\section{Conclusion}
\vspace{-0.4cm}
In this work, we propose \sysn as a way to make synthetic data generation more interpretable. To improve the interpretability, we train models with five different acquisition functions as reward functions using GRPO, and use the resulting \sysn models to generate data for student models. We make two major findings. Firstly, student models trained on \sysn data achieve good performance in both in-distribution and out-of-distribution settings, achieving average performance gains of 4\% compared to the baselines. Secondly, we see that \sysn data can generalize and be used to train models of different architectures and of different sizes. Additionally, it can be used for any kind of training paradigm (ICL, SFT, and RL), which broadens the practical applicability of our method to settings where fine-tuning may not be feasible. To summarize the tradeoffs between acquisition rewards, Confidence is the best general-purpose reward, and Diversity offers the best cost-to-OOD-gain tradeoff. Overall, our work shows that using acquisition rewards to train data generators teaches models what good data is and how to optimally generate good data. In the future, we hope to further refine \sysn to be a Universal \sysn model: one model to generate data for all domains, with one unified notion of data utility. This can help us create a self-evolving loop where models generate data for themselves, teach themselves skills, and automatically update their own curriculum.
\paragraph{Future Work} Our work serves as a proof of concept that data generation models can be optimized, and introduces a framework for evaluating them. Future directions include: (1) combining acquisition rewards linearly or polynomially, (2) optimizing the \sysn training data mixture, and (3) testing our framework on non-verifiable tasks such as instruction following.


\bibliographystyle{plainnat}
\bibliography{custom}

\appendix

\section{The Upper Bound: using GPT as a labeler}
\label{sec: gpt_answer}

In this appendix, we introduce RQ5: how do models trained with \sysn data compare to models trained with data distilled from proprietary models?

\input{results_tables/gpt_results}
We break this experiment into two questions: (a) how accurate are the answers generated from the \sysn models? and (b) what would be the difference if the answers were instead generated by oracle, how accurate would student models be? This experiment is to showcase the upper bound of performance we can get from student models trained with \sysn models. To test (a), we ask GPT-4o-mini whether the generated answer from \sysn is correct or not. According to the ``\% Correct'' column in Table \ref{tab: gpt_labels}, an average of 52.4\% of answers are incorrect. To answer how this affects the answer of the student model (b), we use the questions generated by \sysn and answer them using GPT-4o-mini, and train the student model again (similar to setup for RQ1). According to the ``GPT as Labeler'' column in Table \ref{tab: gpt_labels}, student models achieve an average 6.6\% boost in performance with more accurate labels.

\section{Details of \sysn training}
\label{app: prompts}

In this section, we elaborate on the \sysn training. We use 500 training samples from our selected benchmark during GRPO training. We give these 500 training samples as in-context learning samples. The reason why we use 500 samples is to avoid reward hacking. We saw that with 2,000 training samples, the trained \sysn model generated the exact same training sample every time -- this is because the model reward-hacked, and found one sample that maximized the acquisition reward (or at least, one peak of it). We see that 500 samples is a good balance between learning to maximize the acquisition reward and reward hacking.

Our training prompts for Numina and MedMCQA are in Figures \ref{prompt: numina} and \ref{prompt: medmcqa}, respectively.

\begin{figure}[h]
  \centering
  \begin{tcolorbox}[
    colback=gray!5!white,
    colframe=black!75!black,
    title=Prompt for \sysn training on Numina,
    boxrule=0.3mm,
    width=\textwidth,
    arc=1.5mm,
    auto outer arc
  ]

\begin{verbatim}
Generate a NEW, ORIGINAL math problem that is AS DIFFICULT than the 
reference below. Do NOT copy, paraphrase, or reuse it in any way.

Reference (difficulty calibration only — do not reproduce):
<question>
{question}
</question>

<reasoning>
{reasoning}
</reasoning>

<answer>
{answer}
</answer>

Requirements for your generated problem:
- Requires non-trivial reasoning steps (no single-step shortcuts)
- Draws from: number theory, combinatorics, algebra, geometry, or 
    probability
- Is self-contained and precisely stated
- Reasoning includes a complete step-by-step derivation
- Answer includes just the final result

IMPORTANT: generate a (question, reasoning, answer) triplet; wrap 
your question, reasoning, and answer in the following special tokens:
<question> Insert your question here. </question>
<reasoning> Insert the thinking and general reasoning here. </reasoning>
<answer> Insert your short answer here. </answer>
\end{verbatim}

  \end{tcolorbox}
  \caption{Prompt used during \sysn training. As input, the prompt takes a \texttt{question}, \texttt{answer}, and \texttt{reasoning} from the Numina training dataset.}
  \label{prompt: numina}
\end{figure}

\begin{figure}[h]
  \centering
  \begin{tcolorbox}[
    colback=gray!5!white,
    colframe=black!75!black,
    title=Prompt for \sysn training on Numina,
    boxrule=0.3mm,
    width=\textwidth,
    arc=1.5mm,
    auto outer arc
  ]

\begin{verbatim}
Generate a NEW, ORIGINAL medical multiple-choice question that is AS 
DIFFICULT as the reference below. Do NOT copy, paraphrase, or reuse 
it in any way.

Reference (difficulty calibration only — do not reproduce):
<question>
{question}
</question>

<reasoning>
{reasoning}
</reasoning>

<answer>
{answer}
</answer>

Requirements for your generated problem:
- Requires non-trivial clinical or biomedical reasoning (no single-step 
    lookups)
- Draws from: pharmacology, pathophysiology, clinical medicine, anatomy,
    or biochemistry
- Includes exactly 4 answer choices labeled A, B, C, D — only one i
    correct
- Plausible distractors that reflect common misconceptions or 
    near-miss diagnoses
- Is self-contained and precisely stated
- Reasoning includes a complete step-by-step clinical justification for
    the correct choice and why distractors are wrong
- Answer includes the correct letter and the full text of the chosen 
    option

IMPORTANT: generate a (question, reasoning, answer) triplet; wrap 
your question, reasoning, and answer in the following special tokens:
<question> Insert your question and answer choices 
    (A/B/C/D) here. </question>
<reasoning> Insert the step-by-step clinical reasoning here, including 
    why each distractor is incorrect. </reasoning>
<answer> Insert the correct letter and its full answer text, e.g. 
    "A: Metformin". </answer>
\end{verbatim}

  \end{tcolorbox}
  \caption{Prompt used during \sysn training. As input, the prompt takes a \texttt{question}, \texttt{answer}, and \texttt{reasoning} from the MedMCQA training dataset.}
  \label{prompt: medmcqa}
\end{figure}



\end{document}

%% file: results_tables/main_results.tex
\begin{table}[h]
\centering
\resizebox{\textwidth}{!}{%
    \begin{tabular}{lr|ccc|ccc|ccc}
    \toprule
                                   &                        & \multicolumn{3}{c}{\textbf{Qwen 3B}} & \multicolumn{3}{c}{\textbf{Llama 8B}} & \multicolumn{3}{c}{\textbf{Qwen 7B}}\\ 
    \cmidrule{3-5} \cmidrule{6-8} \cmidrule{9-11}
    \textbf{Category}                       & \textbf{Method}                 & \textbf{Numina} & \textbf{MedMCQA} & \textbf{Avg.} & \textbf{Numina} & \textbf{MedMCQA} & \textbf{Avg.} & \textbf{Numina} & \textbf{MedMCQA} & \textbf{Avg.} \\ 
    \midrule
                                   & Untrained model        & 28.3  & 55.1   & 41.7 & 32.2 & 75.8 & 54.0 & 29.3 & 67.1 & 48.2\\ 
    \midrule
    \multirow{2}{*}{Selection}     
                                   & Random                 & 28.8 \pos{0.5} & 56.4 \pos{1.3} & 42.6 \pos{0.9} & 43.7 \pos{11.5} & 82.4 \pos{6.6} & 63.0 \pos{9.0} & 22.9 \negi{6.4} & 55.3 \negi{11.8} & 37.9 \negi{10.3} \\
                                   & Filtered               & 29.2 \pos{0.9} & 54.3 \negi{0.8} & 41.7 \pos{0.0} & 45.2 \pos{13.0} & 79.3 \pos{3.5} & 62.2 \pos{8.2} & 34.8 \pos{5.5} & 75.4 \pos{8.3} & 55.1 \pos{6.9} \\ 
    \midrule
    \multirow{3}{*}{Synthesis}     
                                   & Original               & 29.4 \pos{1.1} & 55.5 \pos{0.4} & 42.4 \pos{0.7} & 44.5 \pos{12.3} & 78.8 \pos{3.0} & 61.6 \pos{7.6} & 36.5 \pos{7.2} & 75.3 \pos{8.2} & 55.9 \pos{7.7} \\
                                   & DataEnvGym             & 32.3 \pos{4.0} & 55.1 \negi{0.0} & 43.7 \pos{2.0} & 26.8 \negi{5.4} & 44.4 \negi{31.4} & 35.6 \negi{18.4} & 17.8 \negi{11.5} & 76.2 \pos{9.1} & 47.0 \negi{1.2}\\
                                   & Prismatic Syn.         & 44.6 \pos{16.3} & 65.1 \pos{10.0} & 54.8 \pos{13.1} & 33.0 \pos{0.8} & 73.8 \negi{2.0} & 53.4 \negi{0.6} & 34.8 \pos{5.5} & 66.4 \negi{0.7} & 50.6 \pos{2.4} \\ 
    \midrule
    \multirow{6}{*}{\textbf{\sysnshort}} 
                                   & Format                 & 29.6 \pos{1.3} & 57.3 \pos{2.2} & 43.4 \pos{1.7} & 43.1 \pos{10.9} & 81.9 \pos{6.1} & 62.5 \pos{8.5} & 43.2 \pos{13.9} & 74.1 \pos{7.0} & 58.7 \pos{10.45}\\
                                   & Confidence             & 33.7 \pos{5.4} & 59.5 \pos{4.4} & 46.6 \pos{4.9} & 45.5 \pos{13.3} & 81.0 \pos{5.2} & 63.2 \pos{9.2} & 43.1 \pos{13.8} & 75.0 \pos{7.9} & 59.1 \pos{10.9}\\
                                   & Proximity              & 22.1 \negi{6.2} & 63.6 \pos{8.5} & 42.9 \pos{1.2} & 40.7 \pos{8.5} & 77.0 \pos{1.2} & 58.9 \pos{4.9} & 40.2 \pos{10.9}& 74.5 \pos{7.4} & 57.4 \pos{9.2} \\
                                   & Gradient Norm          & 29.7 \pos{1.4} & 58.6 \pos{3.5} & 44.2 \pos{2.5} & 43.7 \pos{11.5} & 80.9 \pos{5.1} & 62.3 \pos{8.3} & 42.5 \pos{13.2} & 70.5 \pos{3.4} & 56.5 \pos{8.3}\\
                                   & Diversity              & 24.7 \negi{3.6} & 58.0 \pos{2.9} & 41.4 \negi{0.3} & 42.8 \pos{10.6} & 77.8 \pos{2.0} & 60.3 \pos{6.3} & 45.2 \pos{15.9} & 75.2 \pos{8.1} & 60.2 \pos{12.0}\\
                                   & Answer Variance        & 30.5 \pos{2.2} & 55.1 \negi{0.0} & 42.8 \pos{1.1} & 43.2 \pos{11.0} & 80.3 \pos{4.5} & 61.7 \pos{7.7} & 43.4 \pos{14.1} & 75.3 \pos{8.2} & 59.4 \pos{11.2} \\
    \bottomrule
    \end{tabular}
}
\vspace{4pt}
\caption{\textbf{In-distribution performance.} Absolute gains/losses are computed relative to the untrained model for each model and benchmark. (\pos{}) and (\negi{}) denote absolute gains and losses relative to the untrained model, respectively.}
\label{tab: main_results}
\end{table}

%% file: results_tables/generalization_qwen.tex
\begin{table}[h]
\centering
\resizebox{\textwidth}{!}{%
\begin{tabular}{lr|cccc|cccc}
\toprule
& & \multicolumn{4}{c|}{\textbf{Trained on Numina}} & \multicolumn{4}{c}{\textbf{Trained on MedMCQA}}\\ 
\cmidrule{3-6} \cmidrule{7-10} 
& \textbf{Evaluated on}           & \textbf{AIME} & \textbf{MedMCQA} & \textbf{CodeForces} & \textbf{Avg.} & \textbf{PubMedQA} & \textbf{Numina} & \textbf{CodeForces} & \textbf{Avg.} \\ 
\midrule
                                     & Qwen 3B         & 43.3 & 55.1 & 51.5 & 50.0 & 28.5 & 28.3 & 51.5 & 36.1 \\ 
\midrule
\multirow{2}{*}{Selection}           
                                     & Random          & 56.7 \pos{13.3} & 62.6 \pos{7.4} & 77.6 \pos{26.1} & 65.6 \pos{15.6} & 30.8 \pos{2.3} & 34.3 \pos{6.0} & 46.9 \negi{4.6} & 37.3 \pos{1.2} \\
                                     & Filtered        & 40.0 \negi{3.3} & 54.3 \negi{0.8} & 78.3 \pos{26.8} & 57.5 \pos{7.5} & 12.5 \negi{16.0} & 29.5 \pos{1.2} & 23.9 \negi{27.6} & 22.0 \negi{14.1} \\ 
\midrule
\multirow{3}{*}{Synthesis}           
                                     & Original        & 40.0 \negi{3.3} & 62.1 \pos{7.0} & 40.1 \negi{11.4} & 47.4 \negi{2.6} & 31.3 \pos{2.8} & 40.0 \pos{11.7} & 40.0 \negi{11.5} & 37.1 \pos{1.0} \\
                                     & DataEnvGym      & 30.0 \negi{13.3} & 65.3 \pos{10.2} & 41.5 \negi{10.0} & 45.6 \negi{4.4} & 45.5 \pos{17.0} & 35.7 \pos{7.4} & 55.7 \pos{4.2} & 45.6 \pos{9.5} \\
                                     & Prismatic Syn.  & 33.3 \negi{10.0} & 60.9 \pos{5.8} & 23.6 \negi{27.9} & 39.3 \negi{10.7} & 28.2 \negi{0.3} & 48.1 \pos{19.8} & 30.9 \negi{20.6} & 35.7 \negi{0.4} \\ 
\midrule
\multirow{6}{*}{\textbf{\sysnshort}} 
                                     & Format          & 40.0 \negi{3.3} & 58.3 \pos{3.2} & 49.7 \negi{1.8} & 49.4 \negi{0.6} & 26.2 \negi{2.3} & 34.2 \pos{5.9} & 38.5 \negi{13.0} & 33.0 \negi{3.1} \\
                                     & Confidence      & 60.0 \pos{16.7} & 63.1 \pos{8.0} & 63.9 \pos{12.4} & 62.4 \pos{12.4} & 38.5 \pos{10.0} & 38.1 \pos{9.8} & 53.4 \pos{1.9} & 43.3 \pos{7.2} \\
                                     & Proximity       & 26.7 \negi{16.7} & 48.4 \negi{6.8} & 50.5 \negi{1.0} & 41.9 \negi{8.1} & 32.5 \pos{4.0} & 43.2 \pos{14.9} & 54.0 \pos{2.5} & 43.2 \pos{7.1} \\
                                     & Gradient Norm   & 43.3 \pos{0.0} & 60.6 \pos{5.5} & 53.7 \pos{2.2} & 52.6 \pos{2.6} & 32.8 \pos{4.3} & 36.4 \pos{8.1} & 45.8 \negi{5.7} & 38.3 \pos{2.2} \\
                                     & Diversity       & 40.0 \negi{3.3} & 51.2 \negi{3.9} & 70.6 \pos{19.1} & 53.9 \pos{4.0} & 35.8 \pos{7.3} & 38.1 \pos{9.8} & 67.6 \pos{16.1} & 47.1 \pos{11.1} \\
                                     & Answer Variance & 40.0 \negi{3.3} & 61.3 \pos{6.1} & 55.0 \pos{3.5} & 52.1 \pos{2.1} & 30.3 \pos{1.8} & 38.4 \pos{10.1} & 52.3 \pos{0.8} & 40.3 \pos{4.2} \\
\bottomrule
\end{tabular}
}
\vspace{4pt}
\caption{\textbf{Generalization results on Qwen.} This table shows the performance of student models trained on curated data from one distribution, but evaluated using other distributions (on the same domain, or different domains). For \textbf{Numina}, AIME is in the same domain, while MedMCQA and CodeForces are in different domains. Similarly, for \textbf{MedMCQA}, PubMedQA is in the same domain, while Numina and CodeForces are in different domains. (\pos{}) and (\negi{}) indicate absolute improvements and degradations relative to the untrained model, respectively.}
\label{tab: generalization_results_qwen}
\end{table}

%% file: results_tables/generalization_llama.tex
\begin{table}[h]
\centering
\resizebox{\textwidth}{!}{%
\begin{tabular}{lr|cccc|cccc}
\toprule
& & \multicolumn{4}{c|}{\textbf{Trained on Numina}} & \multicolumn{4}{c}{\textbf{Trained on MedMCQA}}\\ 
\cmidrule{3-6} \cmidrule{7-10} 
& \textbf{Evaluated on}           & \textbf{AIME} & \textbf{MedMCQA} & \textbf{CodeForces} & \textbf{Avg.} & \textbf{PubMedQA} & \textbf{Numina} & \textbf{CodeForces} & \textbf{Avg.} \\ 
\midrule
                                     & Llama 8B        & 26.7 & 75.8 & 48.7 & 50.4 & 39.0 & 32.2 & 48.7 & 40.0 \\ 
\midrule
\multirow{2}{*}{Selection}           
                                     & Random          & 63.3 \pos{36.7} & 77.4 \pos{1.5} & 15.0 \negi{33.7} & 51.9 \pos{1.5} & 12.5 \negi{26.5} & 38.2 \pos{6.0} & 19.2 \negi{29.5} & 23.3 \negi{16.7} \\
                                     & Filtered        & 70.0 \pos{43.3} & 82.0 \pos{6.2} & 10.0 \negi{38.7} & 54.0 \pos{3.6} & 30.0 \negi{9.0} & 38.4 \pos{6.2} & 25.0 \negi{23.7} & 31.1 \negi{8.8} \\ 
\midrule
\multirow{3}{*}{Synthesis}           
                                     & Original        & 23.3 \negi{3.3} & 41.9 \negi{33.9} & 47.7 \negi{1.0} & 37.6 \negi{12.7} & 36.5 \negi{2.5} & 27.4 \negi{4.9} & 25.9 \negi{22.8} & 29.9 \negi{10.1} \\
                                     & DataEnvGym      & 23.3 \negi{3.3} & 44.6 \negi{31.3} & 20.6 \negi{28.1} & 29.5 \negi{20.9} & 34.0 \negi{5.0} & 26.7 \negi{5.5} & 20.2 \negi{28.5} & 27.0 \negi{13.0} \\
                                     & Prismatic Syn.  & 13.3 \negi{13.3} & 62.8 \negi{13.1} & 23.6 \negi{25.1} & 33.2 \negi{17.2} & 36.3 \negi{2.8} & 28.4 \negi{3.8} & 32.5 \negi{16.2} & 32.4 \negi{7.6} \\ 
\midrule
\multirow{6}{*}{\textbf{\sysnshort}} 
                                     & Format          & 60.0 \pos{33.3} & 81.9 \pos{6.1} & 45.8 \negi{2.9} & 62.6 \pos{12.2} & 43.8 \pos{4.8} & 36.6 \pos{4.4} & 40.0 \negi{8.7} & 40.1 \pos{0.2} \\
                                     & Confidence      & 30.0 \pos{3.3} & 81.0 \pos{5.2} & 41.3 \negi{7.4} & 50.8 \pos{0.4} & 38.8 \negi{0.3} & 39.4 \pos{7.2} & 27.5 \negi{21.2} & 35.2 \negi{4.8} \\
                                     & Proximity       & 26.7 \pos{0.0} & 77.0 \pos{1.2} & 57.5 \pos{8.8} & 53.7 \pos{3.3} & 40.0 \pos{1.0} & 43.3 \pos{11.1} & 45.7 \negi{3.0} & 43.0 \pos{3.0} \\
                                     & Gradient Norm   & 30.0 \pos{3.3} & 80.9 \pos{5.0} & 31.6 \negi{17.1} & 47.5 \negi{2.9} & 38.0 \negi{1.0} & 38.7 \pos{6.5} & 55.6 \pos{6.9} & 44.1 \pos{4.1} \\
                                     & Diversity       & 40.0 \pos{13.3} & 77.8 \pos{2.0} & 20.5 \negi{28.2} & 46.1 \negi{4.3} & 34.5 \negi{4.5} & 40.4 \pos{8.2} & 35.5 \negi{13.2} & 36.8 \negi{3.2} \\
                                     & Answer Variance & 30.0 \pos{3.3} & 80.3 \pos{4.4} & 32.2 \negi{16.5} & 47.5 \negi{2.9} & 32.8 \negi{6.3} & 38.5 \pos{6.3} & 34.3 \negi{14.4} & 35.2 \negi{4.8} \\
\bottomrule
\end{tabular}
}
\vspace{4pt}
\caption{\textbf{Generalization results on Llama.} It is similar to Table \ref{tab: generalization_results_qwen}. (\pos{}) and (\negi{}) indicate absolute improvements and degradations relative to the untrained model, respectively.}.
\label{tab: generalization_results_llama}
\end{table}

%% file: results_tables/model_transfer.tex
\begin{table}[h]
\centering
\resizebox{\textwidth}{!}{%
\begin{tabular}{lr|cccc|cccc}
\toprule
& \textbf{Llama 8B trained on }    & \multicolumn{4}{c|}{\textbf{Numina}} & \multicolumn{4}{c}{\textbf{MedMCQA}}\\ 
\cmidrule{3-6} \cmidrule{7-10}
& \textbf{Evaluated on}           & \textbf{Llama 3B} & \textbf{Qwen 3B }& \textbf{Qwen 7B} & \textbf{Avg.} & \textbf{Llama 3B} & \textbf{Qwen 3B} & \textbf{Qwen 7B} & \textbf{Avg.} \\ 
\midrule
                                     & Untrained           & 33.57 & 28.27 & 29.27 & 30.37 & 49.03 & 55.15 & 67.13 & 57.10 \\ 
\midrule
\multirow{3}{*}{Synthesis}           
                                     & Original            & 41.85 \pos{8.28} & 41.06 \pos{12.79} & 44.96 \pos{15.69} & 42.62 \pos{12.25} & 65.95 \pos{16.92} & 61.45 \pos{6.30} & 61.84 \negi{5.29} & 63.08 \pos{5.98} \\
                                     & DataEnvGym          & 24.67 \negi{8.90} & 24.48 \negi{3.79} & 28.47 \negi{0.80} & 25.87 \negi{4.50} & 22.96 \negi{26.07} & 49.97 \negi{5.18} & 51.07 \negi{16.06} & 41.33 \negi{15.77} \\
                                     & Prismatic Syn.      & 19.88 \negi{13.69} & 21.87 \negi{6.40} & 42.76 \pos{13.49} & 28.17 \negi{2.20} & 75.60 \pos{26.57} & 60.48 \pos{5.33} & 70.35 \pos{3.22} & 68.81 \pos{11.71} \\ 
\midrule
\multirow{6}{*}{\textbf{\sysnshort}} 
                                     & Format              & 41.36 \pos{7.79} & 45.05 \pos{16.78} & 45.05 \pos{15.78} & 43.82 \pos{13.45} & 68.67 \pos{19.64} & 62.86 \pos{7.71} & 64.37 \negi{2.76} & 65.30 \pos{8.20} \\
                                     & Confidence          & 41.05 \pos{7.48} & 43.26 \pos{14.99} & 47.85 \pos{18.58} & 44.05 \pos{13.68} & 65.62 \pos{16.59} & 57.79 \pos{2.64} & 61.83 \negi{5.30} & 61.75 \pos{4.65} \\
                                     & Proximity           & 36.16 \pos{2.59} & 41.26 \pos{12.99} & 47.25 \pos{17.98} & 41.56 \pos{11.19} & 69.11 \pos{20.08} & 54.57 \negi{0.58} & 69.44 \pos{2.31} & 64.37 \pos{7.27} \\
                                     & Gradient Norm       & 40.26 \pos{6.69} & 42.76 \pos{14.49} & 46.85 \pos{17.58} & 43.29 \pos{12.92} & 64.86 \pos{15.83} & 57.88 \pos{2.73} & 57.59 \negi{9.54} & 60.11 \pos{3.01} \\
                                     & Diversity           & 39.36 \pos{5.79} & 40.76 \pos{12.49} & 47.75 \pos{18.48} & 42.62 \pos{12.25} & 66.06 \pos{17.03} & 59.57 \pos{4.42} & 65.25 \negi{1.88} & 63.63 \pos{6.53} \\
                                     & Answer Variance     & 42.36 \pos{8.79} & 43.56 \pos{15.29} & 44.65 \pos{15.38} & 43.52 \pos{13.15} & 73.65 \pos{24.62} & 58.17 \pos{3.02} & 64.79 \negi{2.34} & 65.54 \pos{8.44} \\
\bottomrule
\end{tabular}
}
\vspace{4pt}
\caption{\textbf{Model transfer results on Llama.} \pos{} and \negi{} indicate absolute improvements and degradations relative to the untrained model, respectively.}
\label{tab: model_transfer_results_llama}
\end{table}

%% file: results_tables/gpt_results.tex
\begin{table}[]
\centering \small
\begin{tabular}{r|cc|cc|cc}
\toprule
& \multicolumn{2}{c|}{SM as Labeler} & \multicolumn{2}{c|}{\% Correct} & \multicolumn{2}{c}{GPT as Labeler} \\ \cmidrule{2-3} \cmidrule{4-5} \cmidrule{6-7}
Dataset       & Numina & MedMCQA & Numina & MedMCQA & Numina & MedMCQA  \\  \midrule
Format          & 29.6   & 57.3    & 27.9   & 58.7    & 37.7 \pos{8.1}            & 61.5 \pos{4.2}             \\
Confidence      & 33.7   & 59.5    & 32.5   & 53.9    & 36.2 \pos{2.5}            & 62.6 \pos{3.1}             \\
Proximity       & 22.1   & 63.6    & 77.0   & 54.3    & 39.8 \pos{17.7}            & 67.1 \pos{3.5}             \\
Gradient Norm   & 29.7   & 58.6    & 29.0   & 58.4    & 38.6 \pos{8.9}            & 63.3 \pos{4.7}             \\
Diversity       & 24.7   & 58.0    & 29.6   & 48.6    & 38.1 \pos{13.4}            & 58.1 \pos{0.1}             \\
Answer Variance & 30.5   & 55.1    & 36.1   & 65.4    & 36.7 \pos{6.2}            & 61.4 \pos{6.3}             \\  \midrule
Average         & 28.4   & 58.7    & 38.7   & 56.6    & 37.9 \pos{9.5}            & 62.3 \pos{3.7}             \\
\bottomrule
\end{tabular}
\caption{The upper bound performance of student models trained on \sysn generated questions. ``SM as Labeler'' is the performance of the student models when the student model (SM) itself is the labeler for the generated questions. These numbers are the same as those in Table \ref{tab: main_results}. ``\% Correct'' refers to the percent of answers generated by the student model that are correct, according to GPT-4o-mini. Finally, ``GPT as Labeler'' is the upper bound performance -- the performance of students models when GPT-4o-mini is the labeler.}
\label{tab: gpt_labels}
\end{table}